\documentclass[a4paper,conference]{IEEEtran}
\IEEEoverridecommandlockouts

\usepackage{cite}
\usepackage{url}

\usepackage{amsmath,amssymb,amsfonts}
\usepackage{algorithmic}
\usepackage{graphicx}
\usepackage{textcomp}
\usepackage{xcolor}
\usepackage[hidelinks]{hyperref}

\usepackage[small]{caption}
\usepackage{graphicx}
\usepackage{amsmath}
\usepackage{booktabs}
\usepackage{microtype}
\usepackage{subcaption}
\usepackage{multirow}
\DeclareMathOperator*{\argmax}{arg\,max} %
\DeclareUnicodeCharacter{2212}{-}
\begin{document}
\title{An Unsupervised Approach towards Varying Human Skin Tone Using Generative Adversarial Networks}

\author{\IEEEauthorblockN{Debapriya Roy, Diganta Mukherjee and Bhabatosh Chanda}
\IEEEauthorblockA{Indian Statistical Institute\\
Kolkata, India\\
Email: debapriyakundu1@gmail.com, diganta@isical.ac.in, chanda@isical.ac.in}
}

\maketitle

\begin{abstract}
With the increasing popularity of augmented and virtual reality, retailers are now focusing more towards customer satisfaction to increase the amount of sales. Although augmented reality is not a new concept but it has gained much needed attention over the past few years. Our present work is targeted towards this direction which may be used to enhance user experience in various virtual and augmented reality based applications. We propose a model to change skin tone of a person. Given any input image of a person or a group of persons with some value indicating the desired change of skin color towards fairness or darkness, this method can change the skin tone of the persons in the image. This is an unsupervised method and also unconstrained in terms of pose, illumination, number of persons in the image etc. The goal of this work is to reduce the time and effort which is generally required for changing the skin tone using existing applications (e.g., Photoshop) by professionals or novice. To establish the efficacy of this method we have compared our result with that of some popular photo editor and also with the result of some existing benchmark method related to human attribute manipulation. Rigorous experiments on different datasets show the effectiveness of this method in terms of synthesizing perceptually convincing outputs.
\end{abstract}
\IEEEpeerreviewmaketitle

\section{Introduction}
Augmented reality (AR) is an interactive experience of a real-world environment where the objects that reside in the real world are enhanced by computer-generated augmentations to it, in order to enhance our experiences~\cite{wikiAR, newsMIT}. Though the concept of AR is not new but with the advent of machine learning and deep learning in computer vision, AR got its much needed push into the mainstream. various retailers like Burberry, 1-800-Flowers.com and ASOS, Nike, Benjamin Moore etc. infuse augmented reality (AR) into their apps to help online customers make more informed purchase decisions~\cite{link3}. For example Nike Fit is an app launched by the big sports retail brand Nike that scans one's foot dimensions using his smartphone's camera and suggests his size of shoe. During 2018 CES (Consumer Technology Association) a dressing room app was launched by GAP, a clothing retail company, where shoppers can select the cloth of their choice along with one of five body types to visualize what an outfit will look like on them. All these shows retailers are now focusing more towards new ways to enhance user's shopping experience. Our work is a new approach towards this direction.
\begin{figure}[htp]
\centering
\includegraphics[width=0.44\textwidth]{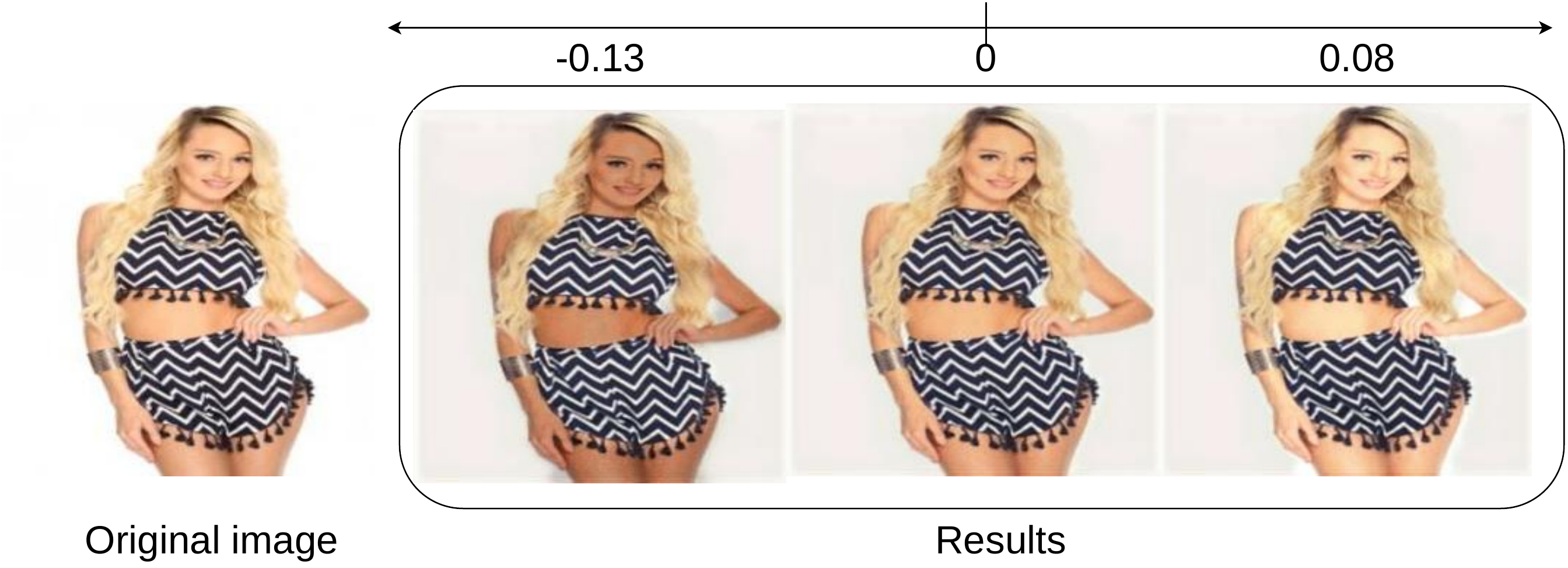}
\caption{Illustrating the objective of the present work. The results with different skin tones is shown in the right ($2^{nd}$ to $4^{th}$). The axis line above indicates the values of the skin tone control variable. The values along the negative direction of the axis indicates darkness while that along the positive direction indicates fairness. The amount of skin color change is proportional to the absolute value of the variable.}
\label{fig: demo}
\end{figure}
During buying cloths online we see the image of a model wearing the cloth and decide to buy or not based on how she looks on it. However excluding the cloth size factor, it is seen often that a confusion remains in the buyer's mind about his looks after wearing it. The reason may be, generally the models are slim, fair skinned, tall etc., which means the seller makes an ideal situation in which his product looks best. Another case may be you are buying a cloth for a friend of yours, however he is much fairer or darker than the model. There can be various such cases where the buyer wishes to see the cloth on a model with his choice of features. Not only clothes, the same concept applies to various accessories also. This work addresses the skin tone aspect of this problem, where we can vary the skin tone of a person in an image. Such an application may be helpful for fashion designers also.

The existing skin tone changing applications are generally targeted to faces. A well known image editing tool is photoshop which is quite popular in this line of work. However these applications require a lot of human intervention and also getting a good result relies heavily on user's expertise and effort. Over the last few years, especially after the advent of generative adversarial network (GAN) face attribute manipulation related various works~\cite{zhang2018sparsely, chen2009automatic, he2019attgan, wang2018weakly, shen2017learning, scfegan} etc., has been proposed. The objective of most of these works are to translate facial images among multiple groups, where each group represents one attribute, e.g., beard, sun-glass, white hair, smile, gender etc. Not only these works are mostly constrained attributes related to facial images but also they use some annotations to divide data into different groups, which requires human intervention. In this work, we attempt an unsupervised approach to change skin tone of person in image. As per our knowledge no previous works has been done to vary the skin color. Although various works has been done on skin detection~\cite{nguyen2018new, kolkur2017human, brand2000comparative, tan2011fusion, zuo2017combining, al2015hybrid, nunez2008detection, kanzawa2011human, shaik2015comparative, kakumanu2007survey, naji2019survey} but the objective of such works are to detect the skin pixels in a given image. However our objective is different in the sense that we not only detect skin pixels but also vary the color of those pixels to synthesize a new image of the input person with varied skin tone (Fig.~\ref{fig: demo}). In fact, the contribution of this paper lies in the idea of changing the skin color. While varying the color of skin pixels might sound like an easy task but the main challenge remains in keeping the perceptual realism in the output, as otherwise the purpose is lost.

We take an adversarial learning~\cite{gan} based approach to solve this problem. We employ a generative adversarial network~\cite{gan} (GAN) to learn a generative model with our proposed skin color distance based loss function. The idea of this loss is to perturb the color of skin pixels based on a control variable, while the amount of perturbation depends on the absolute value of this variable. However, the main challenge here is to vary the skin tone within a permissible range of human skin tones which is addressed by this loss function. Otherwise the realism in the result is lost which is not desirable. To detect the skin / non-skin pixels we employ a skin detector network which is trained in a supervised way. We call our method unsupervised as it does not require any annotation in terms of skin tone.

In summary this paper makes the following contributions,
\begin{itemize}
\item We propose a method to synthesize images of a person over a varying scale of skin tone, where the tone can vary from dark to fair.
\item Inspired from the concept of perceptual loss function we propose a skin color distance based loss function which is used to train a GAN for this current purpose.
\item This is an unsupervised method, hence does not require any skin tone related annotations.
\end{itemize}
As per our knowledge this is the first method to vary skin tone of a person smoothly over a continuous interval. In the rest of the paper a brief literature survey in presented in Sec.~\ref{related_works}, this is followed by our methodology section (Sec.~\ref{methodology}) where we discuss the idea and the workflow of the proposed method in details. To show the effectiveness of this method we presented a detailed qualitative and a quantitative study in Sec.~\ref{experiments}. Finally we conclude in Sec.~\ref{conclusion}.
\section{Related Works}
\label{related_works}
The objective of this work is to smoothly vary human skin tone over a continuous range. Although various works related to attribute manipulation of human has been proposed earlier but the idea of varying skin tone smoothly is relatively new in literature. Below we discuss some of the existing methods related to human attribute manipulation to gain an overview of the existing literature in the current problem context. We also discuss some existing skin / non-skin segmentation methods. %

Earlier methods have explored the ways to manipulate attributes of human faces in images e.g., beard to no-beard looks, or changing the hair color or patterns, smiling to non-smiling face etc. A method for skin beautification has been proposed in~\cite{chen2009automatic} where the authors have used bilateral filter to smooth the detected flawed skin areas and integrate the rectified regions with the original image using Poisson image cloning. After the advent of GAN, attribute manipulation related works have got more attention. Among the recently proposed methods Shen et al.~\cite{shen2017learning} attempted this problem using a GAN based approach. Here the authors employ two generators as image transformation networks and one discriminator. The generators are responsible for the attribute manipulation and its dual operation. The idea of this work is to learn the residual image which refers to the difference of after and before manipulation images. However this method needs labelled data corresponding to each attribute type. A weakly supervised attribute manipulation framework is proposed in~\cite{wang2018weakly}. It employs a GAN which is trained on a perceptual content loss and two adversarial losses to ensure global consistency of the image along with the effect of desired attribute. However this method is weakly-supervised as it learns the attribute from a set of images with that attribute in common. Another method~\cite{zhang2018sparsely} implements a multi-task learning framework based on GAN. It translates images among multiple groups, where each group characterises one attribute. AttGAN~\cite{he2019attgan} proposes a GAN based facial attribute editing framework. It employs an encoder that encodes the input image into its latent representation, which is decoded along with the desired binary attribute vector to synthesize the final image with the desired attributes. A classifier is employed to constrain the attributes of the synthesized image to be as desired. SC-FEGAN~\cite{scfegan} proposed a method that generates images as per user provided inputs e.g., in the form of free-form mask, sketch or color. This is also based on GAN. The generator is an encoder decoder based framework with gated convolutional layers. The speciality of this network is its ability to reconstruct large portions of image erased. The above said methods are all multi-attribute image manipulation works. All of the above methods are either supervised or weakly-supervised requiring some kind of annotation related to the specific attribute. However we attempted the problems of learning human skin tone manipulation in an unsupervised setting where the attribute represented in terms of a value can be varied over a continuous range.

Coming to the problem of human skin, non-skin pixel classification we see earlier in literature various methods has been proposed. Nguyen-Trang et al.~\cite{nguyen2018new} proposed an approach based on Bayesian classifier which detects skin pixels using first posterior probability threshold and "skin candidate" pixels using second posterior probability threshold. Connected component algorithm is used to find all the connected components containing the “skin candidate” pixels. All candidate pixels in a component is classified as skin pixel if it contains at least one skin pixel. Methods based on criterion specific to different color models (e.g., RGB, HSV, YCbCr) has been proposed in~\cite{kolkur2017human, shaik2015comparative}. However colours model based approaches without considering context might face the problem of incorrect classification in some cases e.g., clothing areas with colors close to skin color may be classified as skin. Zuo et al. proposed a deep neural network based approach for skin detection. Where the authors proposed a network integrating recurrent neural networks (RNNs) layers into the fully convolutional neural networks (FCNs). The reason of incorporating RNN layers is to capture semantic contextual dependencies thus overcoming the limitation of convolutional neural networks (CNNs) which captures local features only. In this paper we employed a neural network based skin detection method as a preprocessing step which contributes to our final goal of learning to vary skin color. We do not go into detailed experimental study on our skin detection method as this diverts the main objective of this paper.

\section{Methodology}
\label{methodology}
We propose a method to synthesize a new image corresponding to a given input image with the skin tone of the persons in that image changed according to some desired control variable. Towards this objective we first segment the given image pixels in two classes, skin and non-skin. This is achieved by our skin segmentation network (Sec.~\ref{sec: skin_seg}) which is a Convolutional Neural Network (CNN). The next part of this work which is the final image synthesizer (Sec.~\ref{sec: synthesizer}) employs a Conditional Generative Adversarial Network~\cite{cgan} (cGAN). It takes the image as input, along with the value of a conditional variable and synthesizes a new image with the skin tone of the persons in the image changed in accordance with the value of the variable.
\begin{figure}[h]
\centering
\includegraphics[width=0.44\textwidth]{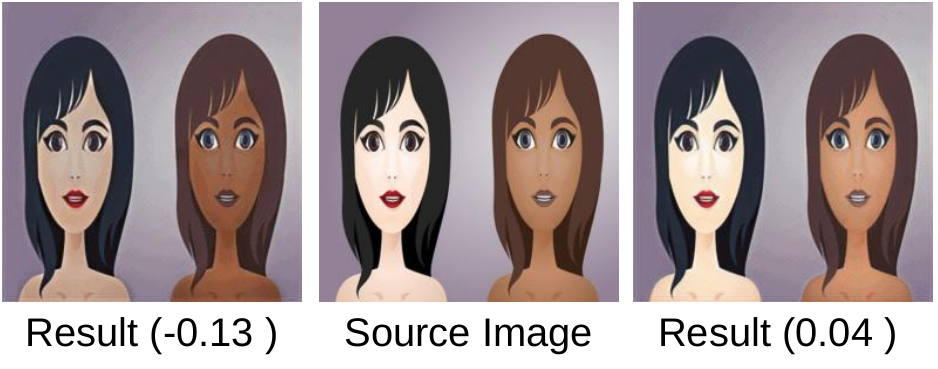}
\caption{Demonstration of change of skin color on image with persons of difference in skin tone. The values within the brackets indicates the value of the skin color control variable $z$. Observe that our method retains the relative difference of skin color.}
\label{fig: res_skin_difference}
\end{figure}
\begin{figure}[h]
\centering
\includegraphics[width=0.44\textwidth]{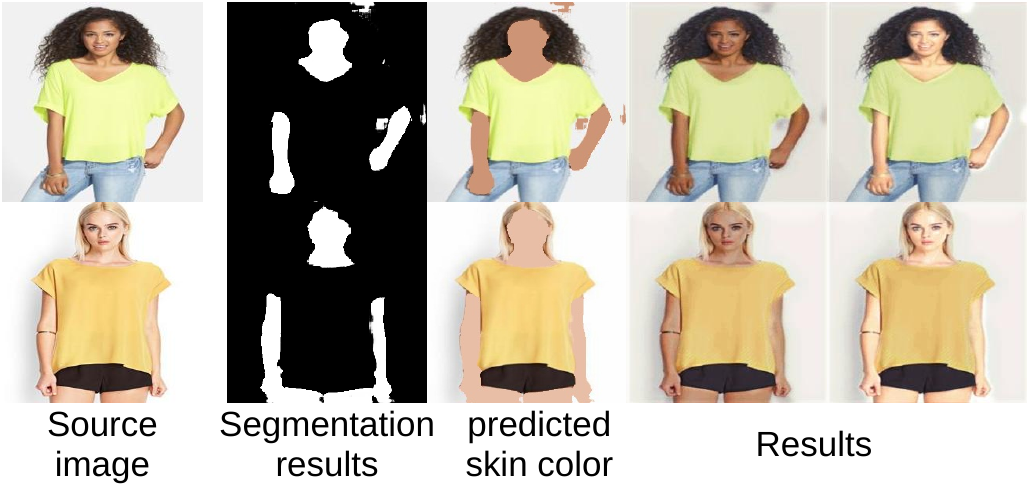}
\caption{Demonstration of results of different stages of the proposed method; ($2^{nd}$ column) Results of skin, non-skin segmentation, ($3^{rd}$ column) demonstration of predicted mean skin tone. For better understanding of the viewer about the prediction accuracy of skin tone the predicted skin color is filled in the skin areas of the source image. ($4^{th}$, $5^{th}$ columns) Results on different values of $z$.}
\label{fig: res_seg_color}
\end{figure}
\subsection{Skin Segmentation}
\label{sec: skin_seg}
The objective of skin segmentation network is two fold. Given an image as input, it first classifies each pixel in the image in two classes $-$ skin, non-skin, followed by estimating a rgb value indicating the skin tone. Here the method predicts one skin tone irrespective of the number of persons present in the image.
During training all the images used contains single person in it therefore no question of confusion arises. However it might be confusing to relate this to images with multiple persons present which might occur during testing. To maintain better flow, we will explain it in detail later.

The skin detection network consists of two sub-networks. The first half of this network is dedicated to skin segmentation objective. This part of the network is trained using the following loss function, 
\begin{equation}
    L_{seg} = L_{c}(\hat{x}_{seg},x_{seg}) + L_{p}(\hat{x}_{seg},x_{seg}) + L_{s}(\hat{x}_{seg},x_{seg}),
\end{equation}
where $L_c$ (. , .) represents count loss and ${L}_{p}$ (. , .), $L_s$ (. , .) represents the perceptual~\cite{perceptual_loss} and SSIM~\cite{ssim} loss respectively between the ground truth (say, $x_{seg}$) and the predicted segmentation result (say, $\hat{x}_{seg}$).

Count loss measures the absolute difference between the counts of skin pixels of the predicted and the ground truth image. Since in general neural networks can not predict binarized output which is required in case of segmentation problems, therefore we try to control this constraint using this loss. It is observed that this loss improves the result in terms of predicting close to binarized output.

SSIM~\cite{ssim} is a popular image metric to measure the structural similarity between two images. However as the value of loss is minimized in neural networks therefore instead of SSIM we employ DSSIM as the loss function which is related to SSIM the following way, DSSIM (. , .) = 1 - SSIM (. , .). Throughout the paper we refer DSSIM loss as SSIM loss.

VGG perceptual loss~\cite{perceptual_loss} is a $L$2 loss between the features of the generated and the groundtruth images, obtained from different layers of pre-trained classification network (VGG-19)~\cite{imagenet}. Instead of exactly matching the pixel values of the generated and groundtruth images this loss matches their feature representations. This encourages the network to produce images which are perceptually similar to their corresponding target images. 

To train the other half of this network which is responsible for the skin color estimation we used MS-SSIM~\cite{ms-ssim} loss, which is a variant of SSIM~\cite{ssim}. As this method is unsupervised therefore we do not use any ground truth skin color annotations for training this network. Instead we fill the detected skin areas with the predicted skin color and minimize the structural dissimilarity between the input and the manipulated image. The reason for designing such objective is to  learn the skin color by maximizing perceptual similarity between the input and the predicted image in an unsupervised way. Some results can be seen in Fig.~\ref{fig: res_seg_color}. The importance of this half of the network lies in learning to extract skin color from the person image itself, which is useful for the next part of the proposed work.

\subsection{Synthesizing Images with Varying Human Skin Tone}
\label{sec: synthesizer}
This section elaborates the main contribution of this work. The problem we are trying to attempt in this section is to synthesize images of varying human skin tone given a source image as input. We formulate this problem as a conditional image generation problem, where the source image, along with its skin segmentation (obtained from the skin segmentation network discussed in the previous section) and a control variable $z$ is given as input to a cGAN. $z$ controls the amount of change of skin tone. The value of $z$ = 0 indices no change of skin color while, values less than zero and above zero indicates the amount of change towards darkness and fairness of skin respectively. Therefore $z$ here plays the role of a skin tone regulator.

We employ a CNN as our generator network in the cGAN and the discriminator network is a patchGAN discriminator~\cite{pix2pix} which is also a CNN. The objective of this generator is to learn a distribution over the input data set. It basically builds a mapping with some conditional information from a prior noise distribution to the data space; while the discriminator predicts the probability of an input given a conditional information, to be coming from the data distribution rather than the generator distribution. In general discriminator predicts a single probability value corresponding to a given input image. However inspired from~\cite{pix2pix} we employ a patchGAN discriminator for this objective which in contrast to classical CNN based discriminator classifies each patch of the image, where the patch size is much smaller than the input image size; which implies pixels separated by more than a patch diameter gets modelled independently. As discussed in~\cite{pix2pix} such a discriminator function acts as texture/style loss and plays significant role in keeping better texture in the generated image.

The interesting part of this cGAN lies in the design of the loss function. Lets denote the generator network as $f_g(.)$, the input image as $x$, $\hat{x}_{seg}$ as the skin segmentation output corresponding to $x$ (we denote complement of $\hat{x}_{seg}$ as ${\hat{x}'}_{seg}$ representing the predicted non-skin regions).
We formulate the objective function in the following way,
\begin{equation}
    L_{cGAN} = l^1 + l^2 + \lambda ( m \times z + l^3 - \epsilon ) + L_{ADV}.
\end{equation}
Where considering $ \hat{x}_{z = 0} = f_{g}(x, z = 0, \hat{x}_{seg})$ and $\hat{x}_{z \neq 0} = f_{g}(x, z \neq 0, \hat{x}_{seg})$, we define, 
\begin{equation}
    \begin{split}
    l^1 &= L_p (\hat{x}_{z = 0}, x)\\
    l^2 &= L_p (\hat{x}_{z = 0} \times {\hat{x}'}_{seg}, x \times {\hat{x}'}_{seg})\\
    l^3 &= \log (0.5 - L_{color})\\
    L_{color} &= L_{p}^{color} (
    \hat{x}_{z \neq 0}, x).
    \end{split}
\end{equation}
Here $\lambda$, $m$ and $\epsilon$ are parameters. $L_{ADV}$ denotes the adversarial loss. The function $L_p(. , .)$ indicates VGG-perceptual loss and $L_{p}^{color}(. , .)$ indicated a loss similar in concept to perceptual loss but the underlying network is the skin color estimation network which is discussed in the previous section. The whole idea behind this loss is to ensure the skin color of the image synthesized by the cGAN with $z$ = 0 is close to that of the original image (this is ensured by the term $l^1$ in the loss function); while that with changing value of $z$ should differ from the original image (ensured by the term $l^3$), preserving the details in regions other than the skin intact (ensured by the term $l^2$). The term $\lambda \times m \times z$ ensures the change in skin color goes in accordance with the value and sign of $z$. During training the cGAN, we sample random values of $z$ from a specified distribution (the noise prior).

Observe from Fig~\ref{fig: res_skin_difference} that this method works in case of images containing people with difference in skin tone. This eradicates the confusion regarding single skin tone prediction by the skin tone estimator sub-network in Sec.~\ref{sec: skin_seg}. We want to clarify that the objective of that part of network was to be able to learn features related to skin color which came into use in training the cGAN. In images containing multiple person, the estimated skin color value basically indicates an average skin tone of all the persons present in the image.
\begin{figure*}[htp]
\centering
\includegraphics[width=0.8\textwidth]{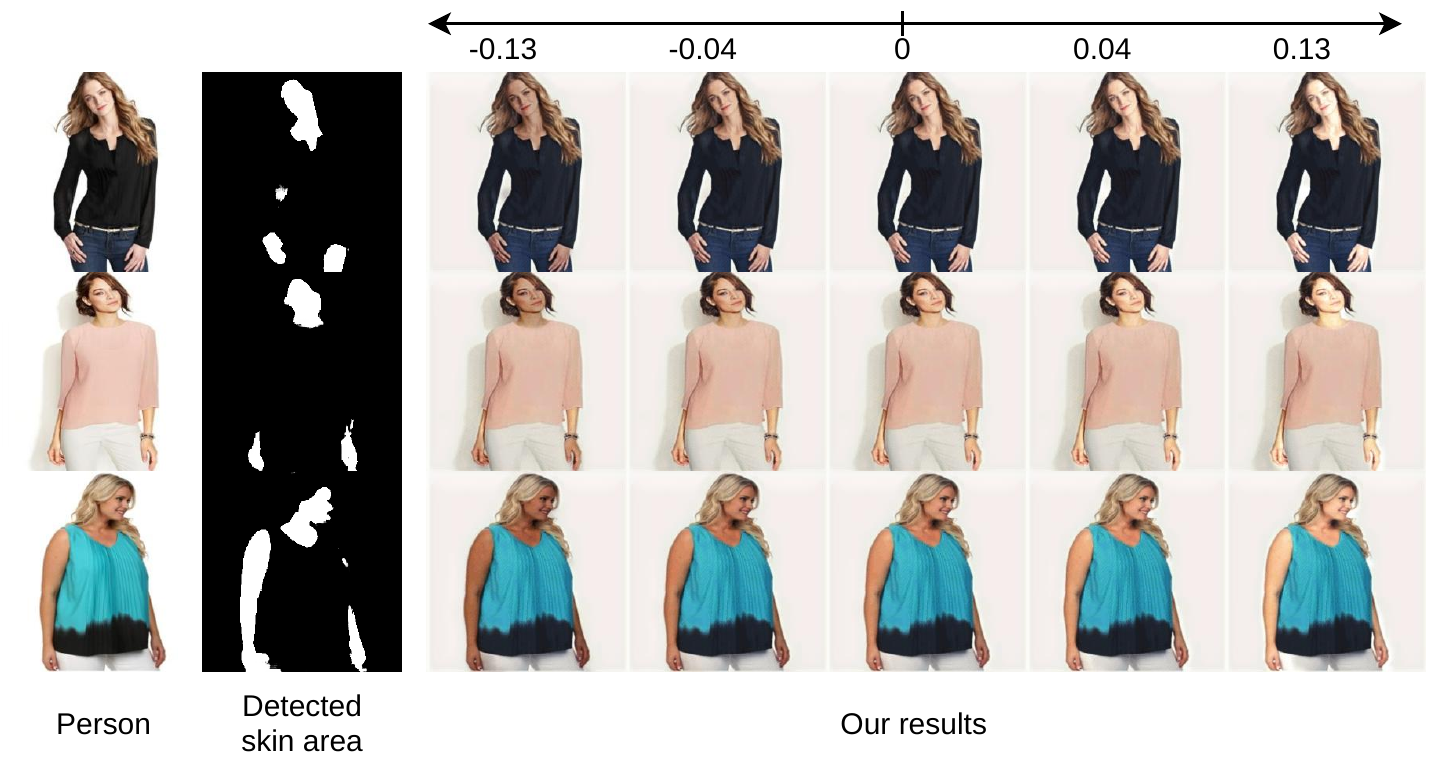}
\caption{Results of our method. The source person images are shown in the 1st column, the next column shows the skin pixel segmentation results. The following columns show the results for different values of $z$. Walking along the axis from negative to positive direction increases the fairness of skin.}
\label{fig: res_final}
\end{figure*}

\section{experiments}
\label{experiments}
\subsection{Dataset}
We evaluate our model based on the following datasets, Category and Attribute Prediction Benchmark, In Shop Cloth Retrieval dataset of DeepFashion~\cite{deepfashion} and MPV~\cite{multiposevton}. DeepFashion is a large scale fashion dataset where images are taken under variety of illumination condition, clothing category, pose and gender.
Deepfashion contains multiple subset of images each with different types of annotations. In this work we used two subsets. 
However our method is applicable to all other subsets of images also. The In Shop Cloth Retrieval dataset contains in total 52,712 images of  multiple views of each person (front, side, back and full) while the Category and Attribute Prediction Benchmark dataset contains 289,222 number of clothing images, where the images are mostly of models wearing the clothing. The MPV dataset contains in total 35,687 images of multiple views of each person. Our model is trained on the Category and Attribute Prediction Benchmark dataset of DeepFashion. For quantitative analysis we tested our model on 2000 randomly selected images from each of the mentioned datasets. For each of the test sample image, the result is obtained
with the values of $z$ drawn from $N$(0 , 0.05).
The qualitative analysis is done on In Shop Cloth Retrieval dataset of DeepFashion, with the cGAN trained on Category and Attribute Prediction Benchmark dataset. For the skin detection part of work we trained the model on LIP~\cite{lipssl} and MPV datasets, where LIP is a popular dataset for human parsing and MPV also contains annotations related to different body parts.
\subsection{Implementation Details}
\subsubsection{Skin Detection Network}
The skin detection sub-network is a hourglass network~\cite{hourglass}, followed by six convolution and 5 pooling layers for the skin color estimation sub-network. The pooling and convolutions are placed alternatively, with number of filters for the convolutions are 18,8,2,3,3,3, with stride 2 and activation relu for all layers except the the last convolution layer. Here the skin detection and skin color estimation sub-networks are two different network with two different purpose, therefore during training they are trained separately. Hourglass network is a CNN (Convolutional Neural Network) that captures features at various scales and is effective for analyzing spatial relationships among different parts of the input. Multiple of these hourglass networks can be stacked together with intermediate supervision for making it deeper. However, for our purpose the stack size of 1 is found to be sufficient. 
\subsubsection{cGAN for Varying Skin Tone}
We have used an hourglass network for the generator and the architecture of patchGAN discriminator is same as it is mentioned in its original paper. The value of $\lambda = 0.003$, $m = 0.004$ and $\epsilon = 0.0002$. These values are chosen experimentally. 

All the networks are trained with optimizer Adam with learning rate 0.0006, $\beta_{1}, \beta_{2}$ = 0.5, 0.999.
 
\subsection{Quantitative Analysis}
\label{quantitative_analysis}
For quantitative analysis we have reported the scores on the following metrics, Inception Score~\cite{is} (IS) and Frechet Inception Distance~\cite{fid} (FID), SSIM~\cite{ssim}. Note that both IS and FID are evaluation measure for GANs, which measures perceptual quality of synthesized outputs, however FID is a more preferred metric~\cite{fid}. We report SSIM score for the training set of images, since groundtruth is not available for test set of images. Here ground truth implies the images of same person in different skin color. We report the value of Kolmogorov-Smirnov test~\cite{kstest} statistic which is a goodness of fit test.
\begin{figure*}[!t]
\centering
\includegraphics[scale=0.9]{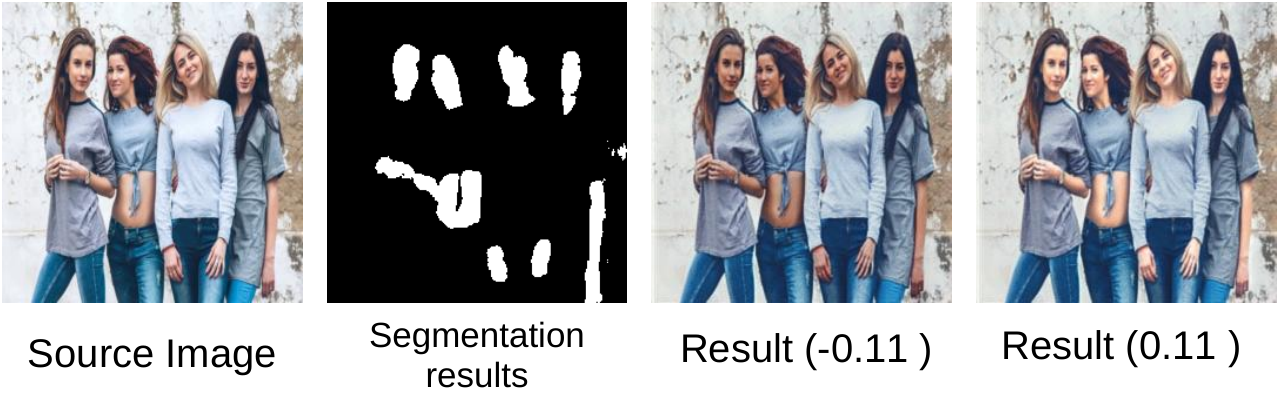}
\caption{Demonstration of result on in-the-wild image (image has been taken from~\cite{image_source0}). The values within the brackets indicates the value of the skin color control variable $z$. Observe that our segmentation module is not constrained to background clutter or presence of multiple persons in image. Also it is worth noticing that the results of the cGAN is perceptually convincing in terms of skin tone variation.}
\label{fig: res_group}
\end{figure*}

\textbf{Inception Score (IS)} measures the classifiability and diverseness in the generated images where the generated images are classified using the inception v3 model~\cite{inceptionv3} to predict the class probability.
IS is defined as,
\begin{equation}{}
    IS (.) = exp (\underset{\textbf{a}\sim p_{g}}{\mathbb{E}}
    [ D_{KL} ( p(b|\textbf{a}) || p(\textbf{a}) )]),
\end{equation}
where \textbf{a} is a generated image sampled from the learned distribution $p_g$, $\mathbb{E}$ is the expectation over the set of generated images, $p(b|\textbf{a})$ is the conditional class ($b$) distribution estimated for image $\textbf{a}$ using Inception model~\cite{inceptionv3} pre-trained with ImageNet~\cite{imagenet}, $p(b)$ = $\underset{\textbf{a}\sim p_{g}}{\mathbb{E}} p(b|\textbf{a})$ is the marginal distribution. $D_{KL}$ is the KL-divergence between the conditional class distribution and the marginal class distribution. IS has a lowest value of 1 and highest value of the number of classes in inception model. Here as we are using inception v3 model which has 1000 number of classes therefore IS has a maximum value of 1000. 

Although IS is observed to correlate well with human perception~\cite{is} but it does not consider real data at all therefore it can not estimate how well the generator approximates the real data distribution. 
To address the issues of IS another metric for evaluating the performance of GAN is proposed. Below we discuss that. 

\textbf{Frechet Inception Distance (FID)} is a measure of similarity between two sets of images. It extracts the features embedded in both real and the generated images from a layer of inception v3 model pre-trained with ImageNet~\cite{imagenet}. Considering the embedding as continuous multivariate Gaussian, the mean and covariance are estimated for both the generated ($\mu_{g}$ , $\sigma_{g}$) and the real data ($\mu_{r}$ , $\sigma_{r}$). Then the FID is calculated as: $\left\lVert\mu_{r} − \mu_{g}\right\rVert_{2}^{2} + Tr( \sigma_{r} + \sigma_{g} − 2 ( \sigma_{r}\sigma_{g} )^{1/2} )$. A lower value of FID is better. 

\textbf{Kolmogorov-Smirnov test (KS test)} is a Goodness-of-fit test that measures the compatibility of random samples against some theoretical probability distribution function. In other words, it is a non-parametric test statistic which defines the largest absolute difference between two cumulative distribution functions as a measure of disagreement. Let us consider, $F_{obs}$ as the empirical distribution function of the data, and $F_{exp}$ as the cumulative distribution function (CDF) associated with the null hypothesis, then the KS test statistic is defined by, 
\begin{equation}
    D_n = \argmax_{} | F_{exp}(.) - F_{obs}(.) |.
\end{equation}
Since KS test statistic is a measure of distance therefore, a lower value of it indices better distribution similarity (Maximum value of this statistic is 1 and minimum is 0). The classical KS test is based on one dimensional data, however since we are dealing in images therefore, we used the 2D variant of the KS test.
\begin{table}[!h]
\centering
\caption{Values of Incepion Score (IS) and Frechet Inception Distance (FID) and SSIM on results of different data sets.}
\begin{tabular}{llll}
\toprule
Dataset & IS$\uparrow$ & FID $\downarrow$& SSIM$\uparrow$\\
\midrule
In Shop & 3.21 $\pm$ 0.17 & 38.33 & 0.93\\
Category-and-Attribute & 3.58 $\pm$ 0.19 & 36.19 & 0.95\\
MPV & 3.03 $\pm$ 0.23 & 42.56 & 0.92 \\
\bottomrule
\end{tabular}
\label{table_is_fid}
\end{table}

\begin{table}[!h]
\centering
\caption{Values of Kolmogov-Smirnov test (KS test) statistic along with the corresponding P-values on results of different data sets.}
\begin{tabular}{lll}
\toprule
Dataset & KS statistic $\downarrow$ & P-Value$\uparrow$\\
\midrule
DeepFashion (Category-and-Attribute) & 0.0249 & 0.5545\\
MPV & 0.0450 & 0.0837\\
\bottomrule
\end{tabular}
\label{table_ks}
\end{table}
We present the values of IS, FID and SSIM in Table.~\ref{table_is_fid} and the values of KS test statistic in Table~\ref{table_ks}. The values of SSIM are based on the results of the cGAN generator with $z$ = 0. The scores of IS and FID, suggests that our method synthesizes quite good quality images which can also be verified visually from the results presented in the qualitative analysis section. Note that values of IS is low in the results. This is because, this problem is related to human images only therefore there is no question of class diverseness in the generated images, where class diverseness relate to the number of classes of InceptionV3 model. The values of KS statistic from Table.~\ref{table_ks} (at 5\% level of significance) shows the generator's distribution is quite similar to the original data distribution, as the difference between the distributions of original and predicted skin tone is statistically insignificant. Which suggests that the predicted skin tones lie within the range of feasible human skin color range.

\subsection{Qualitative Analysis}
\label{qualitative_analysis}
We present our qualitative results in this section. Fig.~\ref{fig: res_seg_color} shows the results of skin segmentation and the estimated skin color. It is observed that the mean predicted color of skin is visually quite similar to the original tone of skin. 
The final results on varying skin tone is shown in Fig.~\ref{fig: res_final}. It is worth noticing that the color of skin varies under different values of the control variable, while the skin becomes fairer as the value $z$ increases, it becomes darker with decreasing values of $z$. 
\begin{figure*}[!t]
\centering
\includegraphics[scale=0.4]{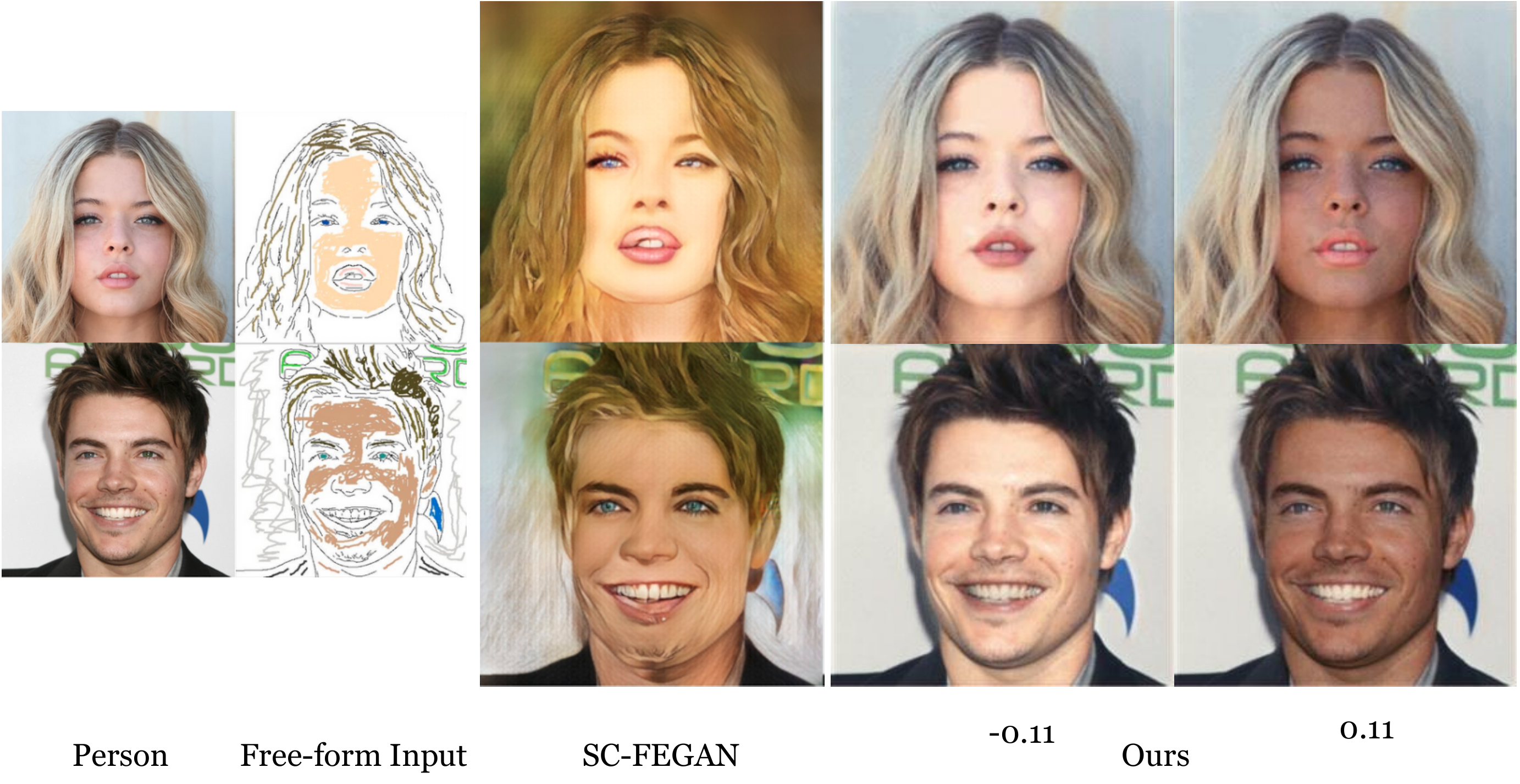}
\caption{Comparative study between the results of our method and that of SC-FEGAN. The results of SC-FEGAN has been taken from the paper~\cite{scfegan} itself. It can be observed that our results look much more realistic compared to that of SC-FEGAN.}
\label{fig: res_compare_scfegan}
\end{figure*}
It is observed experimentally that the value of $z$ within $\pm{0.13}$ gives perceptually coherent images on average. We also present result on in-the-wild image (Fig.~\ref{fig: res_group}). This shows this method is unconstrained to background clutter, variation of skin colors in different persons in the image.

We present a visual comparison with the results of SC-FEGAN~\cite{scfegan} in Fig.~\ref{fig: res_compare_scfegan}. SC-FEGAN is a method for attribute manipulation which can not be directly applied for problems like skin tone change. However this method can reconstruct faces from its free-form input e.g., sketch of the face with color details etc. In Fig.~\ref{fig: res_compare_scfegan} we have presented a visual comparison of our results with the results of SC-FEGAN generated from free-form inputs. Note that in case of our method the idea is to relatively make the skin fairer or darker, hence we do not explicitly provide any skin color as input. Here, for the purpose of better visual comparison we generated two results showing relatively darker and fairer (skin tones) reconstructions of the input faces. It is observed from the figures that our results look visually much convincing in terms of realism compared to that of SC-FEGAN.
\begin{figure}[h]
\centering
\includegraphics[scale=0.45]{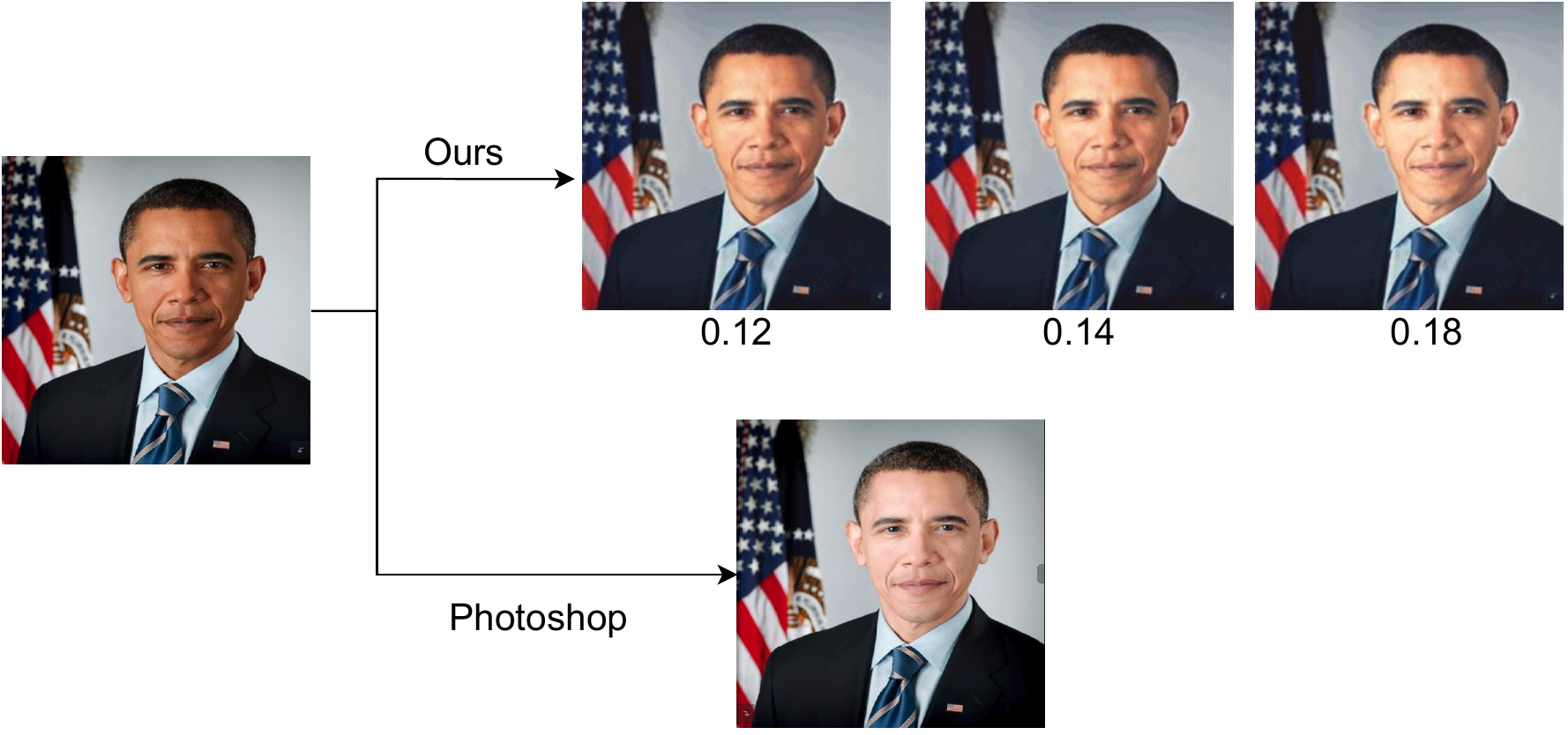}
\caption{Comparative study between the results of our method and that generated by a professional in Photoshop photo editor~\cite{image_source2}. It can be noticed that our method makes consistent gradual change of skin color from its original color producing comparable results.}
\label{fig: res_compare_photoshop}
\end{figure}

In Fig.~\ref{fig: res_compare_photoshop} we have presented a visual comparison between our results with the result generated by a professional in Photoshop photo editor. Here the editor has made the person's skin tone fair. For the purpose of comparison we also synthesized our results towards fairness. Notice the difference between the skin colour in our results and that in the image generated by the editor. This happens because our method changes the tone gradually while keeping coherence with the skin tone in the previous image. Another thing that might be interesting is that as shown in~\cite{video_source2} it took almost 9 minutes to generate the result in photoshop. However compared to this our results can be generated in much less time subject to the deep learning framework and GPU used.

\begin{figure}[h]
\centering
\includegraphics[scale=0.34]{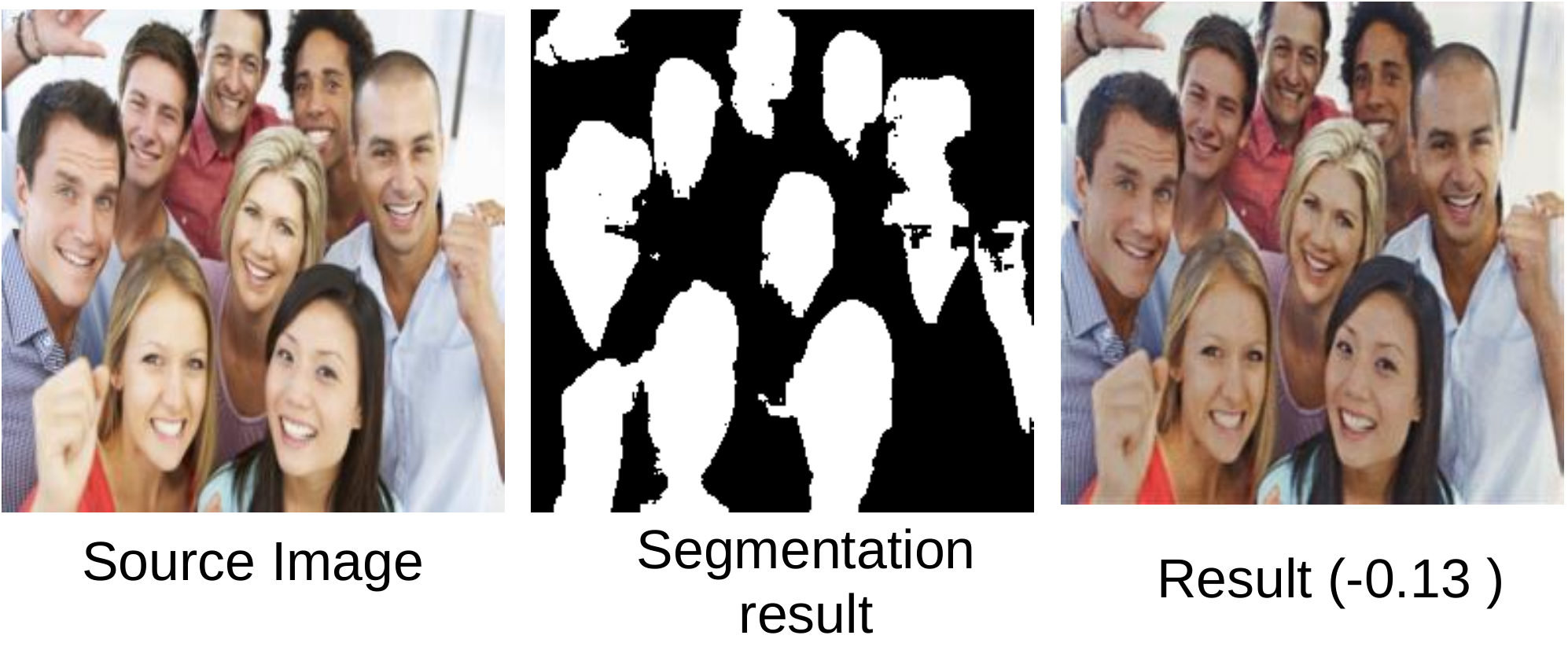}
\caption{Demonstration of limitations of the proposed method (image is taken from~\cite{image_source1}). Improper segmentation has caused uneven skin tone on the leftmost and the rightmost persons.[Please zoom-in for details.]}
\label{fig: res_group_limit}
\end{figure}

Coming to the limitations of this method, we would like to mention that this method is sensitive to irregularities in skin segmentations. As observed from Fig.~\ref{fig: res_group_limit} improper segmentation may result in uneven skin tone (leftmost and rightmost persons).

\section{Conclusion}
\label{conclusion}
This paper presents a method to synthesize new images of a person with varied skin tone, where the amount of variation can be controlled by a control variable. To achieve this objective we first trained a skin segmentation network which segments the skin, non-skin pixels. This is followed by a generative adversarial network which takes as input the source image, along with the skin segmentation result and the value of the corresponding control variable and synthesizes a new image with changed skin color according to the value of the control variable. Experiments on different datasets as well as with the results of popular photo editor and benchmark attribute manipulation related work establishes the usefulness of this method. It is verifiable from both the qualitative and quantitative analysis that this method generates perceptually convincing results.
\bibliographystyle{plain}
\bibliography{bibliography}

\end{document}